\documentclass[lettersize,journal]{IEEEtran}
\usepackage{amsmath,amsfonts}
\usepackage{algorithmic}
\usepackage{algorithm}
\usepackage{array}
\usepackage[caption=false,font=normalsize,labelfont=sf,textfont=sf]{subfig}
\usepackage{textcomp}
\usepackage{stfloats}
\usepackage{url}
\usepackage{verbatim}
\usepackage{graphicx}
\usepackage{cite}

\usepackage{booktabs} 
\usepackage{multirow}
\usepackage{xcolor}
\usepackage{arydshln}
\usepackage{hyperref}
\usepackage{epstopdf}

\newcommand\norm[1]{\Vert#1\Vert}
\newcommand\G{\nabla}

\begin{document}

\title{Normalizing Batch Normalization for Long-Tailed Recognition}

\author{Yuxiang Bao$^{*}$, Guoliang Kang$^{*}$, Linlin Yang, Xiaoyue Duan, Bo Zhao, Baochang Zhang$^{\dagger}$
\thanks{$^{*}$Equal contribution. $^{\dagger}$Corresponding author.} 
\thanks{Yuxiang Bao, Guoliang Kang, Xiaoyue Duan, and Baochang Zhang are with Beihang University, Beijing 100191, China~(e-mail: yxbao@buaa.edu.cn, kgl.prml@gmail.com, LittleMoon@buaa.edu.cn, bczhang@buaa.edu.cn)}
\thanks{Baochang Zhang is also with Hangzhou Research Institute, Beihang, and Nanchang Institute of Technology, Nanchang, China.}
\thanks{Linlin Yang is with Communication University of China~(e-mail: lyang@cuc.edu.cn)}
\thanks{Bo Zhao is with Shanghai Jiao Tong University~(e-mail: bo.zhao@sjtu.edu.cn)}
\thanks{This paper has supplementary downloadable material available at http://ieeexplore.ieee.org., provided by the author. The material includes additional experiments to further demonstrate our method. Contact yxbao@buaa.edu.cn for further questions about this work.}
\thanks{This paper was produced by the IEEE Publication Technology Group. They are in Piscataway, NJ.}
\thanks{Manuscript received 19 December 2023; revised 30 October 2024; accepted 28 November 2024.}}

\markboth{Journal of \LaTeX\ Class Files,~Vol.~14, No.~8, August~2021}%
{Shell \MakeLowercase{\textit{et al.}}: A Sample Article Using IEEEtran.cls for IEEE Journals}

\IEEEpubid{\\\\\\0000--0000/00\$00.00~\copyright~2021 IEEE}

\maketitle

\begin{abstract}
In real-world scenarios, the number of training samples across classes usually subjects to a long-tailed distribution.
The conventionally trained network may achieve unexpected inferior performance on the rare class compared to the frequent class.
Most previous works attempt to rectify the network bias from the data-level or from the classifier-level. 
Differently, in this paper, we identify that the bias towards the frequent class may be encoded into features,  i.e., 
the rare-specific features which play a key role in discriminating the rare class are much weaker than the frequent-specific features. 
Based on such an observation, we introduce a simple yet effective approach, 
normalizing the parameters of Batch Normalization (BN) layer 
to explicitly rectify the feature bias. To achieve this end,   we represent the Weight/Bias parameters of a BN layer as a vector, normalize it into a unit one and multiply the unit vector by a scalar 
learnable parameter. Through decoupling the direction and magnitude of parameters in BN layer to learn, the Weight/Bias exhibits a more balanced distribution and thus the strength of features becomes more even.
Extensive experiments on various long-tailed recognition benchmarks (i.e., CIFAR-10/100-LT, ImageNet-LT and iNaturalist 2018) show that our method outperforms previous state-of-the-arts remarkably.
The code and checkpoints are available at \href{https://github.com/yuxiangbao/NBN}{https://github.com/yuxiangbao/NBN}.
\end{abstract}


\section{Introduction}
\IEEEPARstart{R}{ecent} years have witnessed a rising of computer vision applications, which greatly benefits from large-scale datasets, 
\emph{e.g.}, ImageNet~\cite{ILSVRC15}, COCO~\cite{lin2014microsoft}. These datasets are usually balanced, \emph{i.e.}, the number of samples across different categories are comparable. 
However, in real world, the distribution of the number of samples across different classes may be long-tailed,
which means the number of samples from ``frequent'' classes may be much larger than those from ``rare'' classes.
Consequently, the conventional training process may be biased towards frequent classes, 
making the accuracy on the rare classes not satisfactory.
As in practice we usually expect the model to achieve equally good performance on each class, 
a series of recognition methods ~\cite{cao2019learning,cui2019class,kang2019decoupling,alshammari2022long} are designed for the long-tailed setting, aiming to deal with the above issue.

\begin{figure}[t]
	\includegraphics[width=\linewidth]{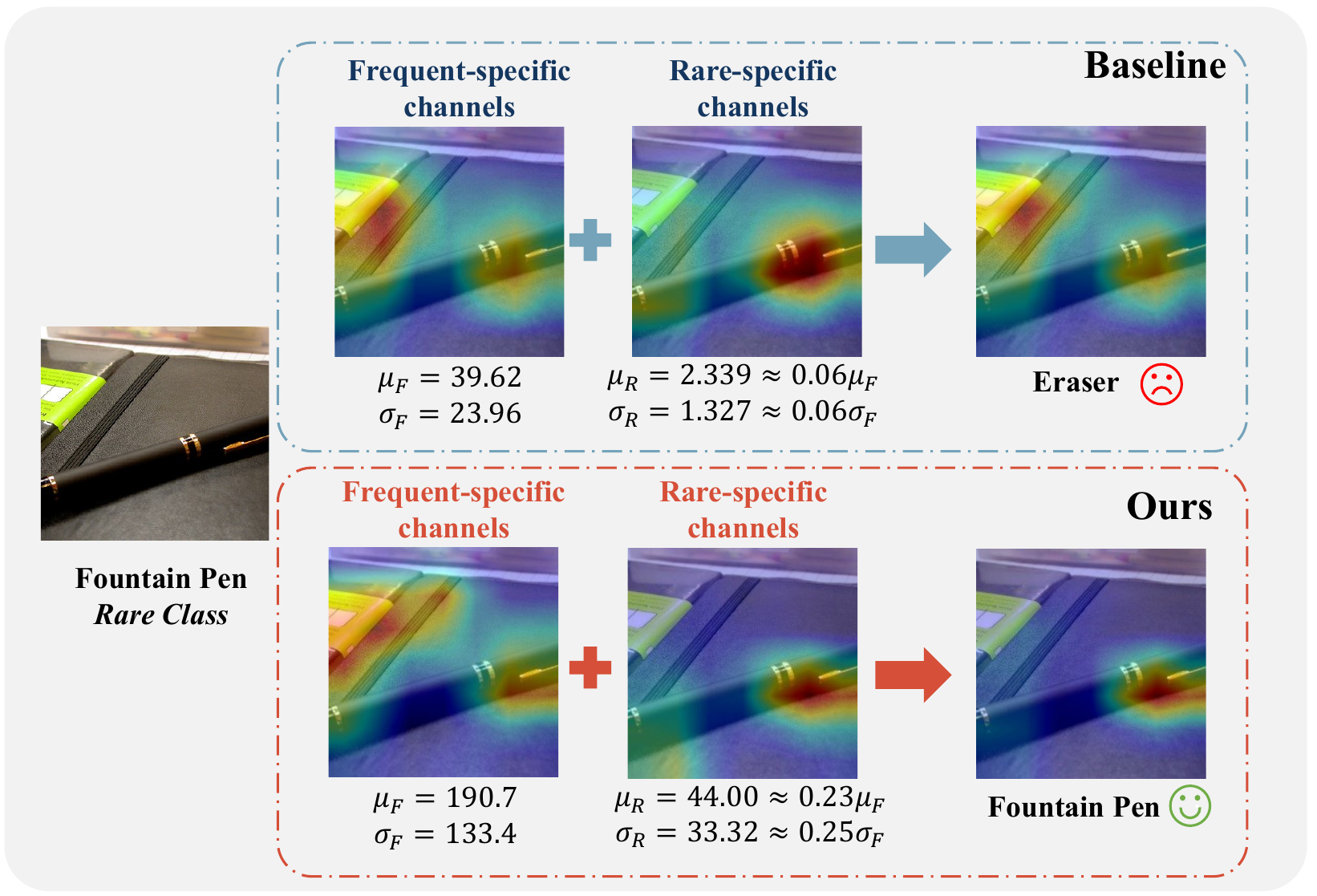}
    \vspace{-4mm}
	\caption{Illustration of feature bias with a rare class sample. We visualize the attention maps~\cite{zagoruyko2016paying} with respect to feature channels which are uniquely important for scoring ``Fountain Pen" (the second column of attention maps) and ``Eraser" (the first column of attention maps) respectively.  Besides, we visualize the attention maps of all feature channels. The statistics of feature channels, including mean $\mu$, and standard deviation $\sigma$, are calculated with samples in the balanced test set to represent the strength of features. For baseline, though the rare-specific features emerge, their weak strength makes the overall attention unexpectedly focus on irrelevant regions. In contrast, our method strengthens those rare-specific features, enabling the model to focus on class-discriminative regions.
    See more details in our supplementary materials.}
    \label{Fig: motivation}
\vspace{-6mm}
\end {figure}

The key of long-tailed recognition problem lies in how to rectify the network bias towards the frequent classes. 
Previous works mainly rectify the bias either from the data-level (\emph{e.g.}, data rebalancing~\cite{drummond2003c4, japkowicz2002class}, data augmentation~\cite{chawla2002smote, park2022majority}) 
or the classifier-level (classifier-decouple training~\cite{kang2019decoupling}, loss function designs~\cite{tan2020equalization, ren2020balanced}).
Despite remarkable progress achieved by those works, they don't explicitly rectify the bias in features which may largely determine the generalization ability of models. Consequently, the feature bias resulted by long-tailed distribution cannot be effectively reduced and 
thus largely restricts the improvement of long-tailed recognition performance.

\IEEEpubidadjcol
From a high-level, different feature channels may represent different characteristics of the image. 
We may evaluate the importance of a feature channel to a specific class via the classifier weight connecting corresponding feature and class.
Feature channels which are important for both classes indicates the characteristics which are shared by both classes.
As the characteristics shared by two classes may not help to distinguish samples,
we visualize the feature channels which are uniquely important for scoring one specific class. 
As shown in Fig.~\ref{Fig: motivation}, the ``Fountain Pen" is a rare class while the ``Eraser" is a frequent class. Given an image of ``Fountain Pen", the model incorrectly discriminates it into the ``Eraser" class. 
To understand what features the model relies on to distinguish a ``Fountain Pen" from an ``Eraser", 
we visualize attentions ~\cite{zagoruyko2016paying} of feature channels which are uniquely important for scoring ``Fountain Pen" (the second column of attention maps),
and those which are uniquely important for scoring ``Eraser" (the first column of attention maps).
We also visualize the attention of all feature channels which largely correlates with the final classification.
We observe that the features that uniquely represent ``Fountain Pen" do exist, 
but they may not dominate the classification result as the strength of them are significantly weaker than the features that uniquely represent the ``Eraser".
That means, in conventional training, the features which are important for correctly classifying samples into rare classes are largely under-estimated.
Thus, the overall distribution of the strength of features is strongly biased towards the frequent class (see Fig.~\ref{fig: histogram}),
rendering it hard to make correct classifications on the rare class.

In this paper, we propose to rectify the feature bias via a simple operation, \emph{i.e.}, normalizing the parameters of Batch Normalization layer. As we know, Batch Normalization (BN) layer is a \textit{de facto} recipe for the conventional deep convolutional networks. It typically follows the convolutional layer or fully-connected layer to adjust the statistics of features to reduce the covariate shift. Forwarding through BN, each feature can be roughly viewed as a Gaussian variable with the mean $\beta$ (the Bias parameter of BN) and the standard deviation $\gamma$ (the Weight parameter of BN). 
Larger Weight and Bias means the feature channel is statistically stronger.
Thus, one possible way to rectify the feature bias (\emph{i.e.,}, to make the strengths of different feature channels more balanced distributed) is to make the distribution of Weight and Bias parameters of a BN layer more balanced.
Our strategy is simple: we treat the Weight parameters in one BN layer as a vector, normalize such a vector into a unit one, and multiply the unit vector by a scalar learnable parameter. 
For the Bias parameters of BN layer, we operate in the same way.
In this manner, we decouple the direction and magnitude of the BN Weight/Bias vector to learn, which encourages more balanced distribution 
of mean and variance of features (see Fig.~\ref{Fig: motivation}) 
and thus rectifies the feature bias.

We conduct experiments on various long-tailed recognition benchmarks including CIFAR-10/100-LT~\cite{krizhevsky2009learning, cao2019learning}, ImageNet-LT~\cite{liu2019large} and iNaturalist 2018~\cite{van2018inaturalist}.  
The experiment results show that our method performs favourably against previous state-of-the-arts.
In contrast to some previous methods which are carefully designed to strike a balance between the accuracy of frequent and rare classes, 
we empirically find our method remarkably improves the accuracy of rare classes without sacrificing the accuracy of frequent classes. 
We also demonstrate that our operation is plug-and-play and can be combined with other state-of-the-art approaches (\emph{e.g.}, classifier-bias rectification methods) to improve the accuracy further. 

In a word, our contributions are summarized as follows
\begin{itemize}
\item 
We observe that without explicit rectification, the learned features may be strongly biased towards the frequent classes and its distribution may exhibit unexpected imbalance, \emph{i.e.}, the features, which are specifically important for discriminating rare classes, are weak. 
Consequently, the statistically strong frequent features (which are representatives for the frequent class but not for the rare class) may drive the model to classify the sample into the frequent class, despite that the sample belongs to a rare one. 
\item 
Based on our observations, we introduce a simple yet effective strategy to explicitly rectify the feature bias in the long-tailed scenario, \emph{i.e.}, normalizing the parameters of BN layer. Through decoupling the direction and magnitude of parameters of the BN layer, our method yields more balanced parameters of the BN layer and thus reduces the bias in features.

\item
We extensively demonstrate the effectiveness of our method on various representative long-tailed image recognition benchmarks,  
\emph{i.e.}, Long-tailed CIFAR-10/100~\cite{cao2019learning}, ImageNet-LT~\cite{liu2019large} and iNaturalist 2018~\cite{van2018inaturalist}. 
We demonstrate that our method is plug-and-play and brings consistent improvements compared to previous methods, achieving new state-of-the-arts.
Ablations and analyses verify the necessity and effectiveness of each design.

\end{itemize}

\section{Related Work}

\noindent \textbf{Long-Tailed Recognition.}
Existing strategies for long-tailed recognition can be roughly categorized into three groups: data balancing~\cite{kubat1997addressing, chawla2002smote, he2009learning, kang2019decoupling, park2022majority, lin2017focal}, loss function design~\cite{cao2019learning, tan2020equalization, ren2020balanced, zhang2021distribution}, and model ensembling~\cite{wang2020long, li2022trustworthy}. Data balancing can be further divided into data re-sampling and data re-weighting. It is intuitive to re-sample the input data to ensure the balance of a training batch.  Existing data re-sampling methods either over-sample the rare cases~\cite{van2007experimental, ando2017deep, chu2020feature, li2021metasaug} or generate the tail samples~\cite{zhong2021improving, park2022majority} or using data augmentation techniques~\cite{zhang2017mixup, devries2017improved, zhong2018camstyle, yang2020small}, like mixup~\cite{zhang2017mixup} and cutout~\cite{devries2017improved}. 
Existing data re-weighting methods assign weights to different classes~\cite{ cui2019class} or even to each instance~\cite{lin2017focal}, where the weight should be inversely proportional to its frequency in the training set. Instead of data balancing, another line of works is to design specific loss functions to benefit the rare-class classification. Those works usually encourage large margins between rare and frequent classes~\cite{cao2019learning, ren2020balanced, tan2020equalization, li2022long}.   
Recently, model ensemble strategies~\cite{wang2020long,li2022trustworthy}  introduce multi-expert architectures that learn diverse classifiers in parallel and make the final decision by averaging the predictions of all the experts.
As discussed, most previous works relieve the bias from data-level or classifier-level, not explicitly rectifying the bias that exists in the feature. In this paper, we aim to explicitly deal with the feature bias to improve the long-tailed recognition performance.

\noindent \textbf{Batch Normalization.}
Batch normalization (BN)~\cite{ioffe2015batch} has been widely used as a basic normalization technique in the majority of state-of-the-art architectures. Specifically, it normalizes the activation value with the statistics of a mini-batch data to meet the standard normal distribution and then applies a linear transformation to the standardized value. Previous works~\cite{wang2019transferable, li2016revisiting, zhong2021improving, cheng2022compound} have modified the BN statistics to improve the model's domain adaptation or long-tailed learning ability. For example, CBN~\cite{cheng2022compound} designs a novel compound batch normalization strategy to alleviate the long-tailed issue. In this paper, we aim to rectify the feature bias via normalizing BN parameters in the long-tailed recognition setting.

\section{Methods}

\subsection{Batch Normalization Revisiting}

Batch normalization (BN) is one of the basic components in advanced convolutional neural networks. Generally speaking, it adopts 
mini-batch statistics (\emph{i.e.}, mean and standard deviation) to standardize each feature, and then applies an affine transformation with 
learnable parameters, \emph{i.e.}, Weight ($\gamma$) and Bias ($\beta$) to the standardized feature. Formally, the operation of BN can be represented as
\begin{equation}
y_{k}=\gamma_{k}\hat{x}_{k}+\beta_{k} \quad \text{where} \quad \hat{x}_{k}=\frac{x_{k}-\mu_{{k}}}{\sigma_{{k}}},
\label{eq:BN}
\end{equation}
where $x_{k}$ denotes the feature of the $k$-th channel. 
The $\mu_{{k}}$ and $\sigma_{{k}}$ denote the mean and standard deviation of the $k$-th channel feature respectively. 
With $\mu_{{k}}$ and $\sigma_{{k}}$,  the $x_k$ is standardized to $\hat{x}_{k}$ that subjects to a standard
normal distribution, \emph{i.e.}, $\mathcal{N}\left(0,1\right)$.
During training, $\mu_{{k}}$ and $\sigma_{{k}}$ are estimated online with a mini-batch of data, and 
during testing, we usually adopt the moving average of $\mu_{{k}}$ and $\sigma_{{k}}$ estimated at each training iteration to 
perform feature standardization.

To maintain the representation capacity of the network, the affine transformation with learnable parameters (\emph{i.e.}, the Weight vector $\boldsymbol{\gamma}$ and the Bias vector $\boldsymbol{\beta}$) are introduced to recover the statistics of features. In Eq.~\ref{eq:BN}, $\gamma_{k}$ and $\beta_{k}$ denotes the $k$-th elements of $\boldsymbol{\gamma}$ and $\boldsymbol{\beta}$ respectively. Therefore, the $k$-th feature can be roughly viewed as a random variable which subjects to a normal distribution $\mathcal{N}(\beta_{k}, \gamma^2_{k})$.

For long-tailed recognition, $\boldsymbol{\gamma}$ and $\boldsymbol{\beta}$ are updated by samples from both frequent and rare classes. 
Thus, the features, which are representatives for a rare class but not shared by frequent classes, may not be fully trained, \emph{i.e.}, the mean and variance of such ``rare-specific'' features are under-estimated and relatively small.
This characteristic renders the features inherently biased towards frequent classes,
which may degenerate the classification accuracy on rare classes.

\subsection{Normalizing Batch Normalization}
\label{sec: NBN}

As discussed above, the features, which are exclusively important for discriminating rare classes, may be unexpectedly weakened during the long-tailed training. Such weakness of feature can be reflected by the small Weight ($\gamma_k$) and Bias ($\beta_k$) of corresponding BN (as illustrated in Fig.~\ref{Fig: motivation}). Thus, a straightforward way to reduce such bias in features is to impose a regularization term that minimizes the variance of parameters in a BN layer. However, though such a regularization way can encourage more balanced Weight/Bias parameters of a BN layer, it is task-agnostic and may impair the model capacity. In practice, it is hard to strike a balance between fitting the training data and balancing the parameters of BN layer, which yields sub-optimal results.

In this paper, we propose to normalize the parameters of BN layer (NBN) to rectify the feature bias, without modifying the training objective or adding additional regularization term.  

Specifically, we view the Weight/Bias parameters of a BN layer as a vector and decouple the learning of its magnitude and direction. 
Such an operation is formulated as

\begin{equation}
{{\boldsymbol{y}}}={{g}_{\boldsymbol{\gamma} }}\frac{{{\tilde{\boldsymbol{\gamma}} }}}{\Vert\tilde{\boldsymbol{\gamma}}\Vert}\circ{{\hat{\boldsymbol{x}}}}+{{g}_{\boldsymbol{\beta} }}\frac{{{\tilde{\boldsymbol{\beta}}}}}{\Vert\tilde{\boldsymbol{\beta}}\Vert},
\label{eq:BRN}
\end{equation}

where $\boldsymbol{y}=[{{y}_{1}}, {{y}_{2}}, \cdots, {{y}_{K}}]$ and $\hat{\boldsymbol{x}}=[{{\hat{x}}_{1}}, {{\hat{x}}_{2}}, \cdots, {{\hat{x}}_{K}}]$. The Weight and Bias vector are decoupled: 
${g}_{\boldsymbol{\gamma}}$ and ${g}_{\boldsymbol{\beta}}$ are learnable parameters 
which represent the magnitudes, and the $\tilde{\boldsymbol{\gamma}}/\norm{\boldsymbol{\tilde{\gamma}}}$, and $\tilde{\boldsymbol{\beta}}/\norm{\boldsymbol{\tilde{\beta}}}$ are the unit vectors, which represent the directions. 
The $\circ$ denotes the element-wise product.

\textbf{Why can NBN yield more balanced Weight/Bias vector?}
\label{subsec: analyse NBN}
We take the updating of Weight vector of the BN layer as an example to illustrate NBN's balancing effect. 
In the process of back-propagation, the gradient of the loss function $\mathcal{L}$ with respect to ${\tilde{\gamma}}_{k}$ (the $k$-th element of $\tilde{\boldsymbol{\gamma}}$) is
\begin{equation}
    \G_{\tilde{\gamma}_{k}}\mathcal{L}=\frac{g_{\boldsymbol{\gamma}}}{\norm{\tilde{\boldsymbol{\gamma}}}}\left(\G_{{\gamma}_k}\mathcal{L}+\alpha \frac{\tilde{\gamma}_k}{\norm{\tilde{\boldsymbol{\gamma}}}}\right),
\label{eq: derivative function}
\end{equation}
where ${{\nabla }_{{{{\gamma }}_{k}}}}\mathcal{L}$ refers to the gradient with respect to ${\gamma}_{k}$,
and $\alpha=-\G_{g_{\boldsymbol{\gamma}}}\mathcal{L}$ means the negative value of gradient with respect to the magnitude $g_{\boldsymbol{\gamma}}$. 

Through Eq.~\ref{eq: derivative function}, we can analyze the relationship between the gradient with respect to the normalized Weight  $\tilde{\gamma}_k$ (\emph{i.e.}, $\G_{{{\tilde{\gamma} }_{k}}}\mathcal{L}$) and the gradient with respect to the original Weight ${\gamma}_k$ (\emph{i.e.}, $\G_{{{{\gamma}}_{k}}}\mathcal{L}$).
As in Eq.~\ref{eq: derivative function}, $\frac{g_{\boldsymbol{\gamma}}}{\norm{\tilde{\boldsymbol{\gamma}}}}$ is just a scaling factor which is shared across different channels.
Thus, we focus on the term $R=\alpha\frac{\tilde{\gamma}_k}{\norm{\tilde{\boldsymbol{\gamma}}}}$ to analyze the rectification effects on different channels.
Considering different effects of $R$, there are two different patterns for updating the parameters of the BN layer: 
\textbf{(A)} when $\alpha>0$ and \textbf{(B)} when $\alpha<0$. 

In pattern \textbf{(A)}, as $\alpha=-\G_{g_{\boldsymbol{\gamma}}}\mathcal{L}>0$, $g_\gamma$ will be increased. According to the gradient descent update rule \emph{i.e.,} $\tilde{\gamma}_k \leftarrow \tilde{\gamma}_k - \epsilon \G_{\tilde{\gamma}_{k}}\mathcal{L}$ where $\epsilon$ is the learning rate, $R$ generally penalizes the magnitude of $|\tilde{\gamma}_k|$, which means introducing $R$ encourages $|\tilde{\gamma}_k|$ to be smaller.
However, the intensity of penalties varies across channels.
The larger $|\tilde{\gamma}_k|/\norm{\tilde{\boldsymbol{\gamma}}}$ leads to a stronger penalization on $|\tilde{\gamma}_k|$.
Thus, in this pattern, $|\tilde{\gamma}_k|$ is encouraged to be more evenly distributed.

In pattern \textbf{(B)}, as $\alpha=-\G_{g_{\boldsymbol{\gamma}}}\mathcal{L}<0$,  $g_\gamma$ will be decreased.
Meanwhile, as $g_\gamma \frac{|\tilde{\gamma}_k|}{\norm{\tilde{\boldsymbol{\gamma}}}}$ reflects the variance $\sigma_k$ of $k$-th channel, the decreasing of $g_\gamma$ results in the decreasing of $\sigma_k$, which may impair the ability to fit the attributes with large variance (especially for frequent-specific features).
According to the gradient descent update rule, $R$ generally increases the magnitude of $|\tilde{\gamma}_k|$, where larger $|\tilde{\gamma}_k|/\norm{\tilde{\boldsymbol{\gamma}}}$ encourages larger $|\tilde{\gamma}_k|$.
Thus, in pattern \textbf{(B)}, the model tends to increase $|\tilde{\gamma}_k|$ to compensate for the decreasing of $g_\gamma$ to avoid capacity loss and encourage data-fitting.

\begin{figure}[t]
\begin{center}
\includegraphics[width=0.9\columnwidth]{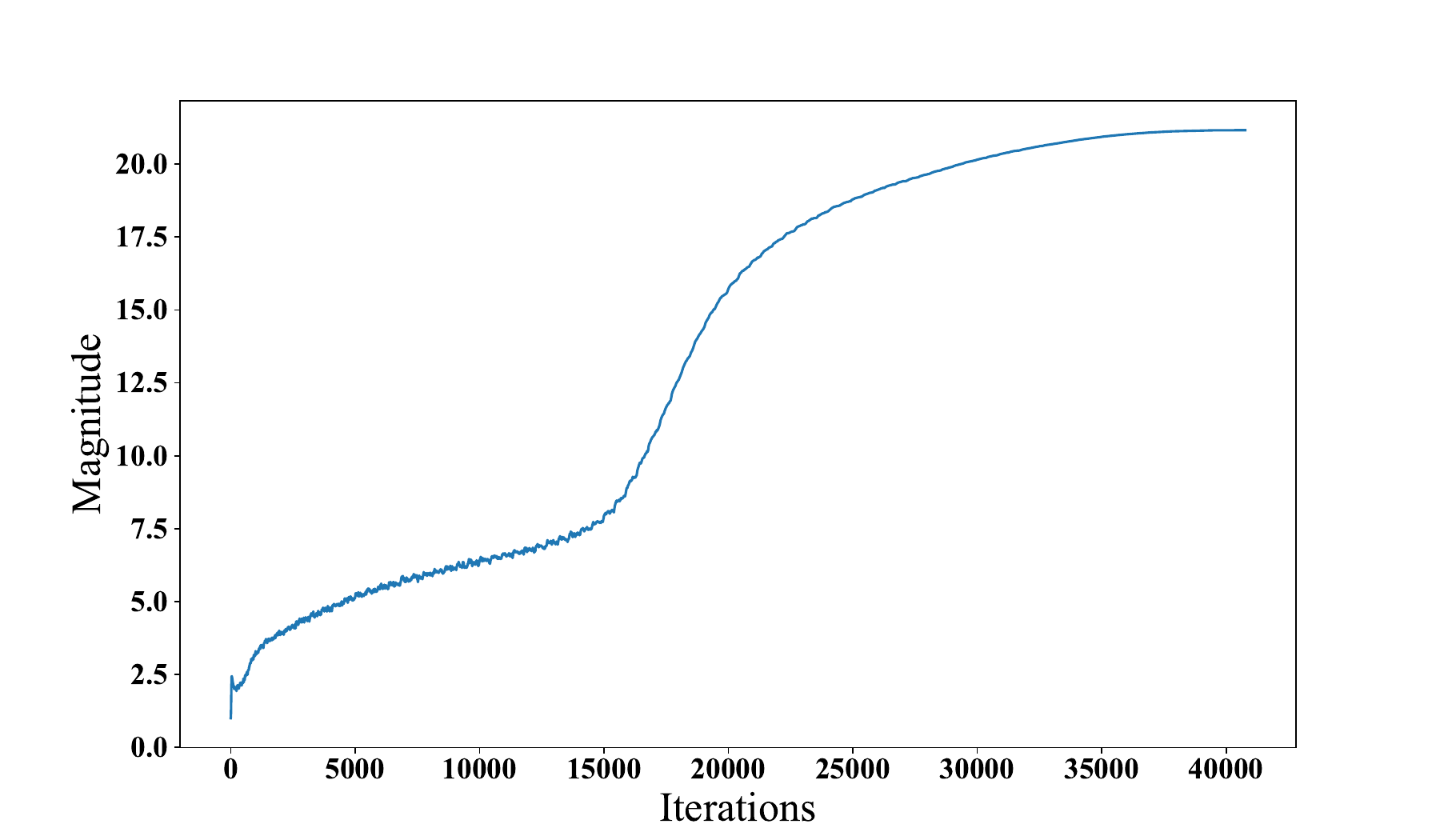}
\caption{Visualization of the magnitude $g_{\boldsymbol{\gamma}}$ during the training process. The results come from experiments on ImageNet-LT. The trends are similar on other datasets.}
\label{Fig:BN weight gradien visualization}
\end{center}
\vspace{-8mm}
\end{figure}

To summarize, the model can adaptively switch between fitting data (pattern \textbf{(B)}) and balancing the parameters of the BN layer (pattern \textbf{(A)}).
As shown in Fig.~\ref{Fig:BN weight gradien visualization}, $g_{\boldsymbol{\gamma}}$ generally increases during training, which means the term $\alpha=-\G_{g_{\boldsymbol{\gamma}}}\mathcal{L}$ usually satisfies $\alpha>0$ (pattern \textbf{(A)}). 
It may be because increasing $g_\gamma$
is much easier and more effective to increase the feature variance $\sigma$. 
Consequently, the parameters of BN become more evenly distributed without disturbing the data-fitting.

\textbf{Connection with Variance Regularization.} 
\label{subsec: NBN vs reg}
As discussed above, one straightforward way to encourage more evenly distributed Weight/Bias parameters of BN layer is to additionally impose
a regularization term that penalizes the variance of Weight/Bias parameters.
In terms of the balancing effect, our NBN shares some similarities with such a variance regularization way.
However, there exist some key differences between NBN and the variance regularization way:
1) Our NBN doesn't modify the training objective, while variance regularization imposes a strong prior which may conflict with the 
training objective. As discussed, NBN can adaptively switch between the data-fitting process and the parameter balancing process. 
As a result, NBN encourages more balanced parameters of BN layer, while not disturbing the fitting to training data.
In contrast, too strong variance regularization may impair the model capacity and degenerate the model's performance.
2) Our NBN doesn't introduce additional hyper-parameters, while for variance regularization, we need to carefully set a hyper-parameter to 
strike a balance between optimizing the training objective and penalizing the variance of parameters.

\textbf{Connection with Weight Normalization~\cite{salimans2016weight}}. 
\label{subsec: connection with WN}
From the operation level, we find that previous Weight Normalization (WN) work~\cite{salimans2016weight} shares a similar idea to ours,
\emph{i.e.}, decoupling the magnitude and direction of convolutional weights to learn.
However, our method is different from WN in nature, as
1) The motivation is different. We aim to rectify the feature bias existing in the long-tailed recognition setting. The NBN is just a simple and effective way to rectify the feature bias in a deep neural network. In contrast, the WN method aims to ease network optimization and accelerate the training. It is originally employed in a network without any BN layers.  
2) The type of layers where the normalization is applied is different and the consequent effect is different. 
Our NBN is applied to the parameters of BN layer, while the WN method is applied to the parameters of convolutional layers.
Thus, our NBN may yield more balanced parameters of BN, while the WN method brings faster convergence. 
Empirically, we find that applying normalization the same way as WN has no effect on improving the long-tailed recognition performance (see Sec.~Ablation Studies).

\begin{figure}[t]
\centering
\includegraphics[width=0.6\linewidth]{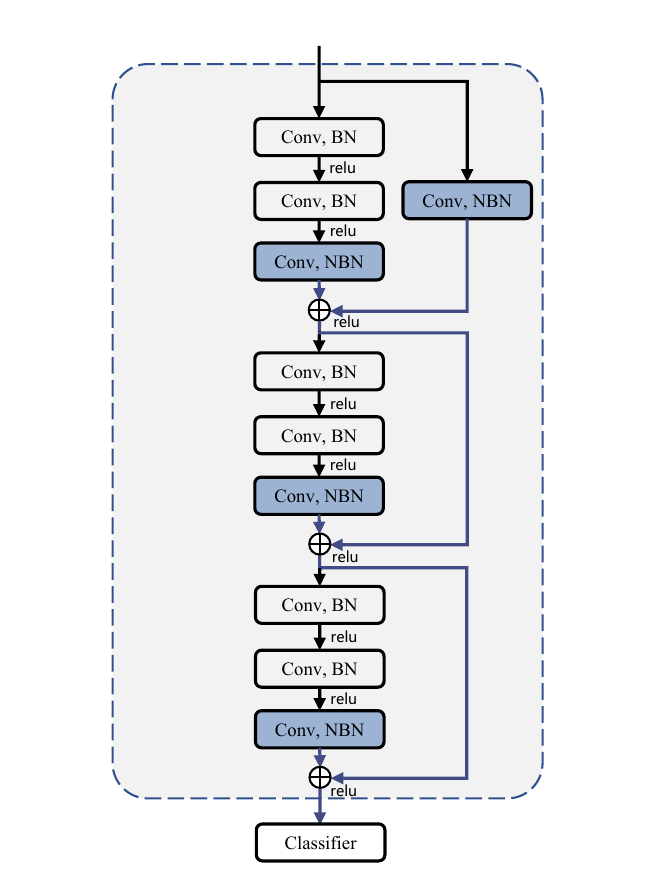}
\vspace{-4mm}
\caption{Illustration of positions to insert the Normalizing Batch Normalization (NBN) layer in the ResNet architecture. We only insert NBN to the last stage of ResNet architecture which consists of three sequential residual blocks.}
\vspace{-4mm}
\label{Fig: method diagram}
\end {figure}

\subsection{NBN in ResNet}
\label{subsec: nbn in resnet}

In this section, we discuss how to insert NBN into the representative ResNet architecture.
We aim to explicitly rectify the bias of features which are the output of the last residual block and directly fed into the classifier to make predictions. 
Meanwhile, through rectification of the last-layer features, we expect that the feature bias from shallow layers may also be
rectified through gradient back-propagation.

As shown in Fig.~\ref{Fig: method diagram}, we replace the original BN layers with our designed NBN layers at specific locations of the last stage of ResNet which consists of three residual blocks.
Through residual connections, the last-layer features are a summation of outputs from three residual blocks. 
Thus, we choose to replace the last BN layers of three residual blocks and 
the BN layer in the first down-sampling residual path with our proposed NBN layers, as illustrated in Fig.~\ref{Fig: method diagram}.
In this way, all sources of bias injected into the last-layer features can be effectively rectified, which yields 
more evenly distributed strength (reflected by corresponding mean and variance) of features.
We empirically verify the effectiveness of our choice in Sec. Ablation Studies.

\subsection{Logits Rectification}
\label{subsec: LR}
In addition to the feature bias, the classifier output also contains bias.
Specifically, the model trained with samples imbalanced distributed across classes tends to allocate higher confidence to the frequent class. 
It is because the statistics (\emph{i.e.}, mean and variance) of logits from frequent classes are inherently larger than 
those from rare classes, rendering the classifier biased towards the frequent classes. 
Thus, we propose to rectify such bias existing in the classifier via a simple Gaussian standardization operation, \emph{i.e.}, 
for each logit, 
\begin{equation}
\hat{z}=\frac{z-\mu}{\sigma},
\label{eq:LogitsBN}
\end{equation}
where $z$ denotes the logit. The $\sigma$ and $\mu$ denote the standard deviation and mean of the logit.
Inspired by BN, during training, we adopt mini-batch statistics of the logit as $\mu$ and $\sigma$,
and for testing, we use the moving average of statistics at each iteration to represent $\mu$ and $\sigma$.

\section{Experiments}

\subsection{Datasets}

We test our method on widely used long-tailed recognition benchmarks including CIFAR-10-LT, CIFAR-100-LT~\cite{krizhevsky2009learning, cao2019learning}, ImageNet-LT~\cite{liu2019large} and iNaturalist 2018~\cite{van2018inaturalist}. 
Following general practice \cite{cui2019class,jamal2020rethinking}, we employ imbalance factor $N_{max}/{N_{min}}$ to depict the extent of imbalance in the specific dataset, where $N_{max}$ and $N_{min}$ denote the number of samples in most and least frequent classes respectively.

\textbf{CIFAR-10/100-LT.} The original CIFAR-10 (CIFAR-100) dataset \cite{krizhevsky2009learning} consists of 50,000 training images and 10,000 test images of size 32 $\times$ 32, which are uniformly distributed among 10 (100) classes. We follow \cite{cui2019class} to manually create the long-tailed version of CIFAR-10 and CIFAR-100 by randomly removing a certain amount of training examples. The imbalance factor of the created dataset is 100. 

\textbf{ImageNet-LT.} We follow \cite{liu2019large} to construct the ImageNet-LT dataset from the original ImageNet dataset \cite{ILSVRC15}. 
The constructed dataset consists of 115.8K training examples belonging to 1,000 classes. The imbalance factor is 256, 
with the numbers of the most and least frequent classes at 1280 and 5 respectively.

\textbf{iNaturalist2018.} The iNaturallist2018 \cite{van2018inaturalist} is real-wold fine-grained visual recognition datasets. It has naturally long-tailed distributions. And it consists of 435,712 training samples within 8,142 classes. The imbalance factor is 500, with the numbers of the most and least frequent classes at 1000 and 2 respectively.

\begin{table}[b]
\caption{Accuracy on CIFAR-10-LT and CIFAR-100-LT with the imbalance factors of 50, 100, and 200.} 
\resizebox{\linewidth}{!}{
\begin{tabular}{@{}lllllll@{}}
\toprule
Dataset                 & \multicolumn{3}{c}{CIFAR-10-LT} & \multicolumn{3}{c}{CIFAR-100-LT} \\ \midrule
\textit{Imbalance Ratio}         & 200      & 100      & 50       & 200       & 100      & 50       \\
Cross Entropy           & 74.0     & 78.0     & 82.3     & 41.2      & 46.1     & 52.0     \\
Focal Loss~\cite{lin2017focal}              & 74.1     & 77.1     & 83.0     & 41.5      & 46.3     & 52.0     \\
LDAM Loss~\cite{cao2019learning}               & 75.0     & 78.9     & 84.3     & 42.3      & 48.3     & 53.3     \\
decoupling LWS~\cite{kang2019decoupling}                     & 78.1     & 83.7     & 84.7     & 47.5      & 53.3     & 58.7     \\
Balanced Softmax~\cite{ren2020balanced}        & 79.8     & 83.6     & 86.5     & 46.8      & 51.7     & 57.6     \\
RIDE~\cite{wang2020long}         & 81.5     & 85.1     & 88.7     & \textbf{51.8}      & 57.3     & 61.7     \\ \midrule
Cross Entropy + Ours    & 77.5     & 81.5     & 86.3     & 46.3      & 50.5     & 56.9     \\
Balanced Softmax + Ours & 80.2     & 84.1     & 87.0     & 50.5      & 53.7     & 58.9     \\
RIDE + Ours  & \textbf{84.5}     & \textbf{87.1}     & \textbf{89.7}     & \textbf{51.8}      & \textbf{57.5}     & \textbf{62.2}     \\ \bottomrule
\end{tabular}}
\label{Tab: CIFAR}
\end{table}

\begin{table}[t]
\caption{
Accuracy on ImageNet-LT.
}
\resizebox{\linewidth}{!}{
\begin{tabular}{@{}ccccc@{}}
\toprule
                        & All  & Tail & Medium & Head \\ \midrule
Cross Entropy      & 45.8 & 9.0  & 39.0   & 67.2 \\
decoupling cRT~\cite{kang2019decoupling}            & 49.6 & 27.4 & 46.2   & 61.8 \\
decoupling LWS~\cite{kang2019decoupling}            & 49.9 & 30.3 & 47.2   & 60.2 \\
Balanced Softmax~\cite{ren2020balanced}   & 51.1 & 30.2 & 48.1   & 62.3 \\
MiSLAS~\cite{cui2021parametric}   & 52.7  & 35.8 & 51.3  & 61.7 \\
PaCo~\cite{cui2021parametric} &56.0 &33.7 &55.7 &64.4 \\
ACE~\cite{cai2021ace} &56.6 &- &- &- \\
ALA~\cite{zhao2022adaptive}  & 52.4  & 35.7  & 49.1  & 62.4 \\
xERM~\cite{zhu2022cross} & 54.1  & 27.5  & 50.0  & \textbf{68.6} \\ 
TSC~\cite{yang2023t}  & 54.2  & -  & -  & - \\
RIDE~\cite{wang2020long}        & 56.8 & 36.0 & 53.8   & 68.2 \\
SuperDisco~\cite{du2023superdisco}  & 57.1	& 37.1	& 53.3	& 66.1 \\
BCL~\cite{zhu2022balanced}  & 56.7 & 36.5 & 53.9   & 67.2 \\
RIDE + MBJ~\cite{liu2022memory}    & 57.7  & 37.7  & 54.1  & 68.4  \\
ABC-Norm~\cite{hsu2023abc}  & 51.7  & 33.1  & 49.7  & 60.7 \\
IIF~\cite{alexandridis2023inverse}  & 52.8  & 18.9  & 47.0  & 72.1 \\
LogN~\cite{zhao2024logit}   & 51.6  & 35.0  & 50.3  & 59.1 \\
CBN~\cite{cheng2022compound}        &57.4   & -  & -  & - \\
\midrule
Cross Entropy + Ours    & 49.4 & 16.2 & 45.2   & 66.4 \\
Balanced Softmax + Ours & 53.8 & 33.4 & 50.9   & 64.6 \\
RIDE + Ours & \textbf{58.2} & 36.7 & \textbf{56.0}   & \textbf{68.6} \\ 
BCL + Ours & 57.7 & \textbf{38.5} & 55.3   & 67.5 \\
\bottomrule
\end{tabular}}
\vspace{-6mm}
\label{Tab: ImageNet_LT}
\end{table}

\subsection{Implementation Details}
\label{subsec:setup}
For CIFAR-10-LT, we adopt ResNet-32~\cite{he2016deep} as our backbone. 
Following \cite{ren2020balanced}, the models are trained for 13,000 iterations with a cosine learning rate schedule, and the warm-up iteration number is set as 800. We adopt the same data augmentation strategies as \cite{ren2020balanced}, including Cutout~\cite{devries2017improved} and auto-augmentation. 
For CIFAR-100-LT, we make some modifications to ResNet-32, making the channels of each layer four times the width of the original version. 
We make such a modification as we empirically find this generally yields better long-tailed recognition performance for both the baseline methods and our method. 
For ImageNet-LT, following \cite{kang2019decoupling, wang2020long, zhu2022balanced}, we utilize ResNeXt-50~\cite{xie2017aggregated} as the backbone in the main comparison and ResNet-50~\cite{he2016deep} for ablation studies. 
The training lasts 90/100 epochs for most of our experiments. 
For the purpose of comparing with the state-of-the-art numbers, 
we cite the number of \cite{du2023superdisco} and \cite{yang2023t} with 200 and 300 training epochs respectively. 
For \cite{cui2021parametric}, we report the result of 180 epochs cited from \cite{zhu2022balanced}. 
For iNaturalist, the experiment setup is similar to ImageNet-LT except that all the experiments are conducted with ResNet-50 trained for 200 epochs.
For the experiments combining our method with the multi-expert model RIDE~\cite{wang2020long}, we follow the same setup as RIDE and adopt its four-expert version for ImageNet-LT and its three-expert version for CIFAR-10/100-LT and iNaturalist.

In all the experiments, we adopt the SGD optimizer with momentum 0.9 to optimize the model.  
We empirically find that sharing the magnitude parameter between the Weight and Bias vector or not achieves comparable results. Thus, for simplicity, we use the same magnitude parameter in NBN, \emph{i.e.}, setting $g=g_\gamma=g_\beta$.
All the models are initialized with random weights. 
For each model, we train it three times and report the average result. 
For a better comparison with previous works, besides the overall accuracy, we additionally report the accuracy with respect to three 
groups~\cite{liu2019large} of classes: Tail group (classes containing less than 20 training samples), Medium group (classes containing 20 to 100 training samples), and Head group (classes containing more than 100 training samples),
on ImageNet-LT and iNaturalist 2018 datasets.

\begin{table}[t]
\caption{Accuracy on iNaturalist 2018. 
}
\resizebox{\linewidth}{!}{
\begin{tabular}{@{}ccccc@{}}
\toprule
                        & All  & Tail & Medium & Head \\ \midrule
Cross Entropy           & 65.2 & 60.0 & 67.2   & 75.8 \\
decoupling cRT~\cite{kang2019decoupling}            & 68.2 & 66.1 & 68.8   & 73.2 \\
Balanced Softmax~\cite{ren2020balanced}        & 69.8 & 69.7 & 70.1   & 69.2 \\
DisAlign~\cite{zhang2021distribution}                & 70.2 & 69.4 & 71.3   & 68.0 \\
MiSLAS~\cite{zhong2021improving}       & 71.6  & 70.4  & 72.4  & 73.2   \\
Balanced Softmax + CMO~\cite{park2022majority}                & 70.9 & 72.3 & 70.0   & 68.8 \\
LTR-WD~\cite{alshammari2022long} & 70.2 & 69.7 & 70.4 & 71.2 \\
GCL~\cite{li2022long}  & 71.0  & 71.5  & 71.3  & 67.5 \\
BCL~\cite{zhu2022balanced}   & 71.7 & 71.6 & 72.0   & 70.9 \\
TSC~\cite{yang2023t}  & 70.3	& -	 & -  & - \\
SuperDisco~\cite{du2023superdisco}  & 73.6	& 71.3	& 72.9	& 72.3  \\
MBJ~\cite{liu2022memory} & 73.2  & -  & -  & -  \\
RIDE~\cite{wang2020long}        & 74.8 & 74.7 & 75.0   & 74.5 \\ 
ABC-Norm\cite{hsu2023abc}  & 71.4  & 70.4  & 73.2  & 68.1  \\
DTRG~\cite{liu2022convolutional}  & 69.5  & -	 & -  & - \\
CBN~\cite{cheng2022compound}  & 74.8   & -	 & -  & - \\
\midrule
Cross Entropy + Ours    & 66.5 & 61.2 & 68.5       & \textbf{76.6}     \\
Balanced Softmax + Ours & 71.6 & 71.3 & 71.7   & 72.1 \\
BCL + Ours & 72.6 & 72.3 & 73.2   & 70.8 \\
RIDE + Ours & \textbf{75.3} & \textbf{75.1} & \textbf{75.4}   & 75.5 \\ 
\bottomrule
\end{tabular}}
\label{Tab: iNat}
\end{table}

\begin{table}[t]
\centering
\caption{Comparison to variance regularization with its strength coefficient $\tilde{\alpha}$ at 0.1 and 1. 
The experiments are conducted on ImageNet-LT and ResNet-50 is adopted.}
\resizebox{\linewidth}{!}{
\begin{tabular}{@{}ccccc@{}}
\toprule
                                                      & All  & Tail & Medium & Head \\ \midrule
Cross Entropy                                              & 42.2 & 6.3  & 34.7   & 64.2 \\
Variance regularization ($\tilde{\alpha}=0.1$) & 44.9 & 10.5 & 38.1   & 65.4 \\
Variance regularization ($\tilde{\alpha}=1$)  & 43.3 & 7.0    & 36.1   & 65.0   \\
NBN                                                   & \textbf{47.3} & \textbf{13.4} & \textbf{40.9}   & \textbf{67.1} \\ \bottomrule
\end{tabular}}
\vspace{-3mm}
\label{Tab: alternative ways}
\end{table}

\subsection{Comparison with State-of-the-Art Methods}
\label{subsec: experiment results}
\textbf{Baselines.} To verify the effectiveness of our method, we make comparisons with previous state-of-the-art methods. The ``Cross Entropy'' means the baseline method that trains the model with cross-entropy loss and doesn't consider the long-tailed distribution of training samples.
The other methods are all the long-tailed recognition methods.
To show our method is plug-and-play, we combine our method with previous typical long-tailed recognition methods, 
\emph{i.e.}, Balanced Softmax~\cite{ren2020balanced} which is a loss function specifically designed for the long-tailed setting, RIDE~\cite{wang2020long} which is an ensemble method with multiple experts, 
and BCL~\cite{zhu2022balanced} which applies supervised contrastive learning to the long-tailed scenario.
To ensure a fair comparison, we run the officially released code of those works under a unified setting and compare them.
We also compare our method with two-stage ones, \emph{i.e.}, ``decoupling cRT'' and ``decoupling LWS''~\cite{kang2019decoupling} which decouple the feature learning and classifier calibration process.

\textbf{Results.} The experiment results are reported in Table~\ref{Tab: CIFAR},~\ref{Tab: ImageNet_LT},~\ref{Tab: iNat} for CIFAR-10/100-LT, ImageNet-LT and iNaturalist, respectively. 
Generally, we have three observations. Firstly, all the long-tailed recognition methods including ours outperform the cross-entropy baseline. For example, our method outperforms the cross-entropy baseline by 3.6\% on ImageNet-LT. 
Secondly, our method can be combined with previous works to improve the accuracy further. 
For example, despite the strong baseline provided by RIDE and BCL, combining with our method yields 
obvious improvement by around 1.4\% and 1.0\% on ImageNet-LT, respectively. 
Last but not least, we observe that it is hard for most of the previous long-tailed recognition methods to strike a balance between the accuracy of ``Head'' group and that of ``Tail'' group.
For example, comparing Balanced Softmax to Cross Entropy on iNaturalist 2018, the accuracy increases by about 10\% for the Tail group, while for the Head group, the accuracy decreases by more than 6\%.
However, in the majority of cases, our method can bring consistent improvements on all the three groups. 

\subsection{Ablation Studies}
\noindent \textbf{Existence of rare-specific and frequent-specific features.}
To clarify the concept of class-specific feature channels, we provide extra experiments to elaborate on it.
To test the different effects of each channel on rare and frequent classes, we mask each feature channel and record
the performance drop with respect to rare classes and frequent classes. If masking a feature channel results in
a severe performance drop for rare classes but a negligible drop for frequent classes, this feature channel is much
more important for rare classes than for frequent classes and can be considered as the ``rare-specific'' channel.
The ``frequent-specific'' channel can be determined in a similar way.
Table~\ref{Tab: mask channels performance} shows the tail and head classes' performance when the rare-specific and frequent-specific channels are masked. 
We observe that different feature channels do have different effects on rare and frequent classes. 
For example, when masking rare-specific channels, the accuracy of tail classes drops heavily (1.0\%), while the accuracy of head classes keeps nearly unchanged.
Besides, the channels which are important for both rare and frequent classes are named ``common" channels. From Table~\ref{Tab: mask channels performance}, we observe that masking those common channels may result in obvious accuracy drop for both tail and head classes.

The existence of rare-specific and frequent-specific features further justifies the motivation of NBN:
Our NBN aims to make $\gamma$ and $\beta$ of BN more evenly distributed across channels.
While $\gamma$ and $\beta$ represent the strength of a feature channel statistically, more evenly distributed $\gamma$ and $\beta$ of BN means the strengths of feature channels are comparable, \emph{i.e.,} the strengths of rare-specific and frequent-specific feature channels are comparable, which may reduce the feature bias and benefit the model's long-tailed recognition performance.

\begin{table}[h]
\centering
\caption{Ablation study on the performance variation when the rare-specific, frequent-specific, and common feature channels are masked, respectively. Experiments are conducted with ResNet-50 on ImageNet-LT.}
\label{Tab: mask channels performance}
\begin{tabular}{@{}cccc@{}}
\toprule
Method                          & \multicolumn{1}{c}{All} & \multicolumn{1}{c}{Tail} & \multicolumn{1}{c}{Head} \\ \midrule
ResNet-50                       & 42.2                    & 6.3                     & 64.2                     \\
mask rare-specific channels     & 41.1                    & 5.3~\footnotesize{(\textcolor{green}{-1.0})}                     & 64.1~\footnotesize{(\textcolor{green}{-0.1})}                     \\
mask frequent-specific channels & 41.4                    & 6.1~\footnotesize{(\textcolor{green}{-0.2})}                     & 62.9~\footnotesize{(\textcolor{green}{-1.3})}                     \\ 
mask common channels  & 37.1    & 3.5~\footnotesize{(\textcolor{green}{-2.8})}     & 58.1~\footnotesize{(\textcolor{green}{-6.1})}   \\
\bottomrule
\end{tabular}
\end{table}

\noindent \textbf{Comparison with Variance Regularization.} 
As shown in Table~\ref{Tab: alternative ways}, we compare our NBN method with the variance regularization which also encourages more balanced parameters of BN layer. It can be seen that both variance regularization and our method outperform the cross-entropy baseline remarkably, which verifies that balancing the parameters of BN layer may reduce the feature bias and thus benefit the long-tailed recognition performance.
We also observe that our NBN shows superior performance than variance regularization. As discussed in Sec.~Methods, it may be because our NBN strikes a balance between balancing the parameters and fitting the data. 
In contrast, stronger variance regularization leads to inferior results, implying variance regularization may conflict with the data-fitting objective.

\begin{table}[t]
\caption{Comparison of different positions to insert NBN. ``Type A'' means we replace all the BN layers in the last stage of ResNet architecture with NBN layer, while ``Type B'' means we employ NBN at positions which are complementary to ``Ours''. ``Type C'' means we replace all BN layers with NBN. The experiments are conducted with ResNet-32 on CIFAR-10 and ResNet-50 on ImageNet-LT.}
\resizebox{\linewidth}{!}{
\begin{tabular}{@{}cccccc@{}}
\toprule
Datasets                     & Normalization type & All  & Tail & Medium & Head \\ \midrule
\multirow{4}{*}{CIFAR-10-LT} & Baseline           & 78.0 & -    & 65.7   & 81.1 \\
                             & Type A             & 80.0 & -    & 69.4   & 82.6 \\
                             & Type B             & 78.6 & -    & 69.3   & 80.9 \\
                             & Type C             & \textbf{80.3} & -    & 68.8 	& \textbf{83.2}  \\
                             & Ours               & 80.0 & -    & \textbf{71.6}   & 82.1 \\ \midrule
\multirow{4}{*}{ImageNet-LT} & Baseline           & 42.2 & 6.3  & 34.7   & 64.2 \\
                             & Type A             & 47.0 & \textbf{13.7} & 40.4   & 66.9 \\
                             & Type B             & 43.0 & 6.8  & 35.8   & 64.9 \\
                             & Type C             & 46.9 & 12.9  & 40.2   & \textbf{67.1} \\
                             & Ours               & \textbf{47.3} & 13.4 & \textbf{40.9}   & \textbf{67.1} \\ \bottomrule
\end{tabular}}
\vspace{-5mm}
\label{Tab:optimal norm strategy}
\end{table}

\begin{table}[t]
\centering
\caption{Effect of sharing $g$ across NBN layers.
Experiments are conducted on ImageNet-LT with ResNet-50.}
\resizebox{0.8\linewidth}{!}{
\begin{tabular}{@{}ccccc@{}}
\toprule
                                     & All  & Tail & Medium & Head \\ \midrule
cross entropy                             & 42.2 & 6.3  & 34.7   & 64.2 \\
not sharing                   & 45.2 & 8.8  & 38.0   & 66.9 \\
partially sharing   & 46.1 & 10.8 & 39.3   & 67.0 \\
sharing all (Ours)  & \textbf{47.3} & \textbf{13.4} & \textbf{40.9}   & \textbf{67.1} \\ \bottomrule
\end{tabular}}
\label{Tab: share g}
\end{table}

\noindent\textbf{Combination with Decoupling cRT.}
We combine our NBN with the prevailing two-stage long-tailed recognition method, \emph{i.e.}, decoupling cRT~\cite{kang2019decoupling}, where the backbone and the classifier are jointly trained with instance-balanced sampling in stage one to learn the visual representation, and in stage two the classifier is calibrated via a re-sampling strategy with the backbone fixed. 
We conduct the experiment on ImageNet-LT~\cite{liu2019large} and iNaturalist 2018~\cite{van2018inaturalist}. As shown in Table~\ref{Tab: two stage}, with the help of NBN rectifying the feature bias, the accuracy is remarkably improved, \emph{e.g.}, around 2\% improvement on ImageNet-LT. Moreover, we empirically find that the learning of magnitude in stage two can further improve the performance by about 1\% on both datasets. 
The results reflect that our method can effectively rectify the bias of features and improve the generalization ability of features beyond instance-balanced sampling.

\begin{table}[t]
\centering
\caption{Accuracy on ImageNet-LT, and iNaturalist of decoupling cRT~\cite{kang2019decoupling} with or without NBN. Following \cite{kang2019decoupling}, in stage one the backbone and the classifier are jointly trained for 90 epochs, while in stage two only the classifier is tuned for 10 or 30 epochs for ImageNet-LT and iNaturalist, respectively. 
The ``NBN'' means we employ our NBN in stage one and keep the parameters of NBN fixed in stage two.
The ``NBN*'' denotes that the magnitude $g$ of NBN is also updated in stage two.}
\resizebox{0.8\linewidth}{!}{
\begin{tabular}{@{}ccccccc@{}}
\toprule
Datasets                     & Stage                    & Methods  & All  & Tail & Medium & Head \\ \midrule

\multirow{5}{*}{ImageNet-LT} & \multirow{2}{*}{Stage 1} & Baseline & 42.2 & 6.3  & 34.7   & 64.2 \\
                             &                          & NBN      & \textbf{47.3} & \textbf{13.4} & \textbf{40.9}   & \textbf{67.1} \\ \cmidrule(l){2-7} 
                             & \multirow{3}{*}{Stage 2} & Baseline & 47.4 & 26.2 & 43.8   & 59.3 \\
                             &                          & NBN      & 49.9 & 24.1 & 46.2   & \textbf{63.5} \\
                             &                          & NBN*     & \textbf{50.7} & \textbf{26.8} & \textbf{47.4}   & 63.3 \\ \midrule
\multirow{5}{*}{iNaturalist} & \multirow{2}{*}{Stage 1} & Baseline & 62.3 & 56.7 & \textbf{64.3}   & \textbf{74.3} \\
                             &                          & NBN      & \textbf{63.3} & \textbf{59.3} & \textbf{64.3}   & 73.4 \\ \cmidrule(l){2-7} 
                             & \multirow{3}{*}{Stage 2} & Baseline & 65.2 & 63.2 & 66.0   & 69.0 \\
                             &                          & NBN      & 65.9 & 63.3 & 66.8   & \textbf{71.4} \\
                             &                          & NBN*     & \textbf{66.9} & \textbf{64.8} & \textbf{67.9}   & 70.4 \\ \bottomrule
\end{tabular}}
\vspace{-3mm}
\label{Tab: two stage}
\end{table}

\noindent \textbf{Effect of different positions to insert NBN.}
In Sec.~Methods, we introduce that we apply NBN to the last stage of ResNet architecture. There are two alternative ways to insert NBN layer. In Table~\ref{Tab:optimal norm strategy}, ``Type A'' means we replace all the BN layers in the last stage with NBN layer, while ``Type B'' means we employ NBN at positions which are complementary to ours. ``Type C'' means we replace all BN layers with NBN.
We observe that ``Type A'' and ``Type C'' achieve comparable results while ``Type B'' is inferior to ours.
All the results show that the positions we select to employ NBN are sufficient (by comparing to ``Type A'' and ``Type C'') and 
necessary (by comparing to ``Type B'') to reduce feature bias. 

\noindent \textbf{Effect of sharing magnitude $g$ across different NBN layers.}
As the feature fed into the classifier is a summation of outputs from the main branch and the residual connections, 
we share the magnitude $g$ across all the NBN layers to ensure these outputs have comparable strength. 
As shown in Table~\ref{Tab: share g}, we compare our way (\emph{i.e.}, ``sharing all'') to two alternative ways:
``not sharing'' which means each NBN layer independently owns one magnitude $g$ and ``partially sharing'' which shares $g$ only within each residual block.
We observe that both ``partially sharing'' and ours perform better than the ``not sharing'' way,
and ours is the best of all.

\noindent \textbf{Effect of learnable magnitude $g$ and normalization.}
To investigate the respective effect of learnable magnitude $g$ and the normalization operation,  
we compare NBN (``Ours'') with the one that fixes $g$ throughout the training process, and the one without normalization operation in Table~\ref{Tab: fix g}.
We observe that fixing $g$ during training performs severely worse than our NBN, and even worse than the cross-entropy baseline, which shows learning $g$ is important for fitting the data to maintain the model capacity. 
Moreover, if we remove normalization from our NBN (``Ours w/o. normalization''), the accuracy remains higher than the cross-entropy baseline but is much worse than NBN, which verifies the effectiveness of performing normalization in NBN.

\begin{table}[]
\centering
\caption{Ablation study on the respective effect of learnable parameters g and decoupling without performing normalization. The experiment is performed with ResNet-50 on ImageNet-Lt.}
\resizebox{\linewidth}{!}{
\begin{tabular}{@{}ccccc@{}}
\toprule
         & Cross Entropy & Ours & Fix $g$ & Ours w/o. normalization \\ \midrule
Acc (\%) & 42.2          & \textbf{47.3} & 17.3          & 44.9                   \\ \bottomrule
\end{tabular}}
\vspace{-2mm}
\label{Tab: fix g}
\end{table}

\noindent \textbf{NBN versus WN~\cite{salimans2016weight}.} 
\label{subsec: normalize other parts}
In Sec.~Methods, we discuss the relationship between NBN and WN. 
From the operation-level, WN is distinct from NBN as WN is applied to the convolutional layers while NBN is applied to BN layers.
In Table~\ref{Tab: ablation1}, we compare NBN with WN. 
We conduct the experiments on CIFAR-10-LT with ResNet-32 and ImageNet-LT with ResNet-50.
For fairness, all the settings are kept the same except for the type of layer to employ the parameter normalization operation.
We observe that WN doesn't achieve obvious improvement compared to the baseline, while our NBN performs remarkably better than WN.
This comparison shows that the original WN cannot deal with the feature bias problem in long-tailed recognition.

\begin{table}[t]
\centering
\caption{Comparisons between NBN and WN~\cite{salimans2016weight}. The experiment is conducted with ResNet-32 on CIFAR-10-LT and ResNet-50 on ImageNet-LT. Cross Entropy denotes training with cross-entropy loss only. }
\resizebox{0.9\linewidth}{!}{
\begin{tabular}{@{}cccccc@{}}
\toprule
Datasets    & Positions                    & All  & Tail & Medium & Head \\ \midrule
CIFAR-10-LT & Cross Entropy                     & 78.0 & -  & 65.7   & 81.1 \\
            & NBN               & \textbf{80.0} & -  & \textbf{69.4}   & \textbf{82.6} \\
            & WN             & 78.3 & -  & 65.5   & 81.5 \\
            \midrule
ImageNet-LT & Cross Entropy                    & 42.2 & 6.3  & 34.7   & 64.2 \\
            & NBN               & \textbf{47.3} & \textbf{13.4} & \textbf{40.9}   & \textbf{67.1} \\
            & WN             & 42.0 & 6.9  & 34.4   & 63.8 \\
            \bottomrule
\end{tabular}
}
\vspace{-5mm}
\label{Tab: ablation1}
\end{table}

\begin{table}[b]
\centering
\vspace{-6mm}
\caption{Performance comparison between cross-entropy baseline and ours on balanced datasets.}
\vspace{-3mm}
\resizebox{0.8\linewidth}{!}{
\begin{tabular}{@{}c|cc|c@{}}
\toprule
Model      & \multicolumn{2}{c|}{ResNet-32} & ResNet-50 \\ \midrule
Dataset    & CIFAR-10      & CIFAR-100      & ImageNet  \\ \midrule
Cross Entropy   & \textbf{93.8}          & 70.5           & 75.3      \\
Ours & 93.4          & 70.5           & \textbf{75.7}      \\ \bottomrule
\end{tabular}}
\label{Tab: balanced datasets}
\end{table}

\noindent \textbf{Normalizing BN versus Normalizing features.}
We have also compared our proposed NBN with normalizing features. Given the feature shape output by BN is [N,C,H,W], we try two normalization ways which normalize features across the dimensions of [C] and [C,H,W] respectively. The results listed in Table~\ref{Tab: feature norm} show that feature normalization along the channel [C] axis also brings performance gain compared to the original training~(about 2\%), but less than that achieved by NBN~(about 5\%).
The difference between NBN and feature normalization is NBN performs normalization in a dataset level (as it normalizes the statistics across channels), while feature normalization performs normalization in a sample level (as it normalizes the features for each sample). We think normalization in a dataset level is more reasonable and effective.

\begin{table}[ht]
\begin{center}
\caption{
Comparison with feature normalization, LN and IN with ResNet-50 on ImageNet-LT.
}
\vspace{-3mm}
\label{Tab: feature norm}
\resizebox{\linewidth}{!}{
\begin{tabular}{@{}ccccc@{}}
\toprule
ImageNet-LT                             & All  & Tail & Medium & Head \\ \midrule
ResNet-50                               & 42.2 & 6.3  & 34.7   & 64.2 \\
ResNet-50 + feature norm on channel {[}C{]}     & 44.5 & 10.4 & 37.8   & 65.0 \\
ResNet-50 + feature norm on channel {[}C,H,W{]} & 39.8 & 0.9  & 28.9   & \textbf{67.1} \\
ResNet-50 + Ours    & \textbf{47.3}  & \textbf{13.4}   & \textbf{40.9}   & \textbf{67.1}  \\
\bottomrule
\end{tabular}}
\vspace{-4mm}
\end{center}
\end{table}

\noindent \textbf{Comparison with LayerNorm and Instance Norm.}
To further demonstrate the superiority of our method, we make comparison with other normalization techniques, LayerNorm~(LN) and InstanceNorm~(IN), under long-tailed recognition setting. 
In our experiment, we replace NBN with LN / IN, or add extra LN / IN layers after the BN layers. 
As demonstrated in Table~\ref{Tab: LN/IN},  compared to the ResNet-50 baseline, replacing BN layers with LN or IN leads to a significant performance drop, while inserting extra LN layers yields improvement~(+1.4\%). However, our NBN achieves more than 5\% performance gain~(43.6\% v.s. 47.3\%). All those results verify the superiority of our NBN compared to LN and IN.

\begin{table}[t]
\caption{Experiments on replacing/adding LN or IN in ResNets.}
\vspace{-3mm}
\label{Tab: LN/IN}
\begin{center}
\begin{tabular}{@{}ccccc@{}}
\toprule
ImageNet-LT                  & All  & Tail & Medium & Head \\ \midrule
ResNet-50                    & 42.2 & 6.3  & 34.7   & 64.2 \\
ResNet-50 (replace BN with LN) & 32.5 & 4.9  & 24.5   & 52.3 \\
ResNet-50 (replace BN with IN) & 27.9 & 0.1  & 15.4   & 53.4 \\
ResNet-50+LN                 & 43.6 & 11.5 & 36.9   & 63.3 \\
ResNet-50+IN                 & 27.8 & 0.2  & 15.1   & 53.3 \\
ResNet-50 + Ours    & \textbf{47.3}  & \textbf{13.4}   & \textbf{40.9}   & \textbf{67.1}  \\ \bottomrule
\end{tabular}
\end{center}
\vspace{-6mm}
\end{table} 

\noindent \textbf{Effect of NBN on LVIS-V1.}
We also extend our method to long-tailed detection and segmentation tasks on LVIS-V1.
All the experiments are conducted with MMDetection and follow a similar training strategy with Seesaw method~\cite{wang2021seesaw} for fair comparison, except that the batch size is set to 4 and the learning rate is set to 0.005 following the linear rule~\cite{goyal2017accurate}. Following \cite{wang2021seesaw}, we adopt a normalized linear classification head for category prediction of proposals. 
For evaluation, we report the detection and segmentation results with mAP, APr, APc, and APf, which are commonly used metrics to evaluate the average performance, performance on rare, common, and frequent groups respectively. 
As shown in Table~\ref{Tab: lvis}, compared to Seesaw, our method increases both the detection and segmentation performance, with $0.7\%$ and $0.6\%$ overall gain respectively. Moreover, considering APr, our method surpasses Seesaw by $1.7\%$ and $1.2\%$. 
The results generally demonstrate our method also benefits the long-tailed detection and segmentation performance.
As our method is not specifically designed for long-tailed detection and segmentation tasks, how to make NBN more suitable for detection and segmentation tasks and improve the performance further remains future work.

\begin{table}[t]
\caption{Results of applying our method to long-tailed detection and segmentation on LVIS-V1.}
\label{Tab: lvis}
\centering
\resizebox{\linewidth}{!}{
\begin{tabular}{@{}ccccccccc@{}}
\toprule
\multirow{2}{*}{2x schedule} & \multicolumn{4}{c}{Detection} & \multicolumn{4}{c}{Segmentation} \\
                             & mAP   & Apr   & Apc   & Apf   & mAP    & Apr    & Apc    & Apf   \\ \midrule
Cross Entropy                & 19.9  & 1.3   & 16.7  & 29.3  & 19.2   & 0.0      & 17.2   & 29.5  \\
EQL                          & 22.0  & 10.7  & 18.8  & 30.6  & 22.7   & 15.1   & 20.4  & 28.6   \\
Seesaw Loss                  & 25.3  & 12.7  & 23.7  & \textbf{32.6}  & 24.7   & 14.9   & 23.9   & 29.8  \\
Cross Entropy + Ours         & 23.5  & 9.2   & 21.0  & 32.4  & 22.5   & 10.3   & 21.0   & 29.6  \\
Seesaw Loss + Ours           & \textbf{26.0}    & \textbf{14.4}  & \textbf{24.5}  & \textbf{32.6}  & \textbf{25.3}   & \textbf{16.1}   & \textbf{25.8}   & \textbf{29.9}  \\ \bottomrule
\end{tabular}}
\vspace{-5mm}
\end{table}

\noindent \textbf{Effect of NBN with balanced dataset.}
Technically, our NBN can also be adopted when the training set is balanced.
As shown in Table~\ref{Tab: balanced datasets}, we employ NBN on original CIFAR-10, CIFAR-100, and ImageNet datasets which are balanced.
We observe that when the dataset is balanced, our method achieves comparable results to the cross-entropy baseline. 
It is because our method is designed to mitigate the feature bias caused by the long-tailed distribution of training samples, 
and when the dataset is balanced, such feature bias doesn't exist, rendering our method ineffective.

\begin{figure*}[t]
\begin{center}
\vspace{-7mm}
\includegraphics[width=\linewidth]{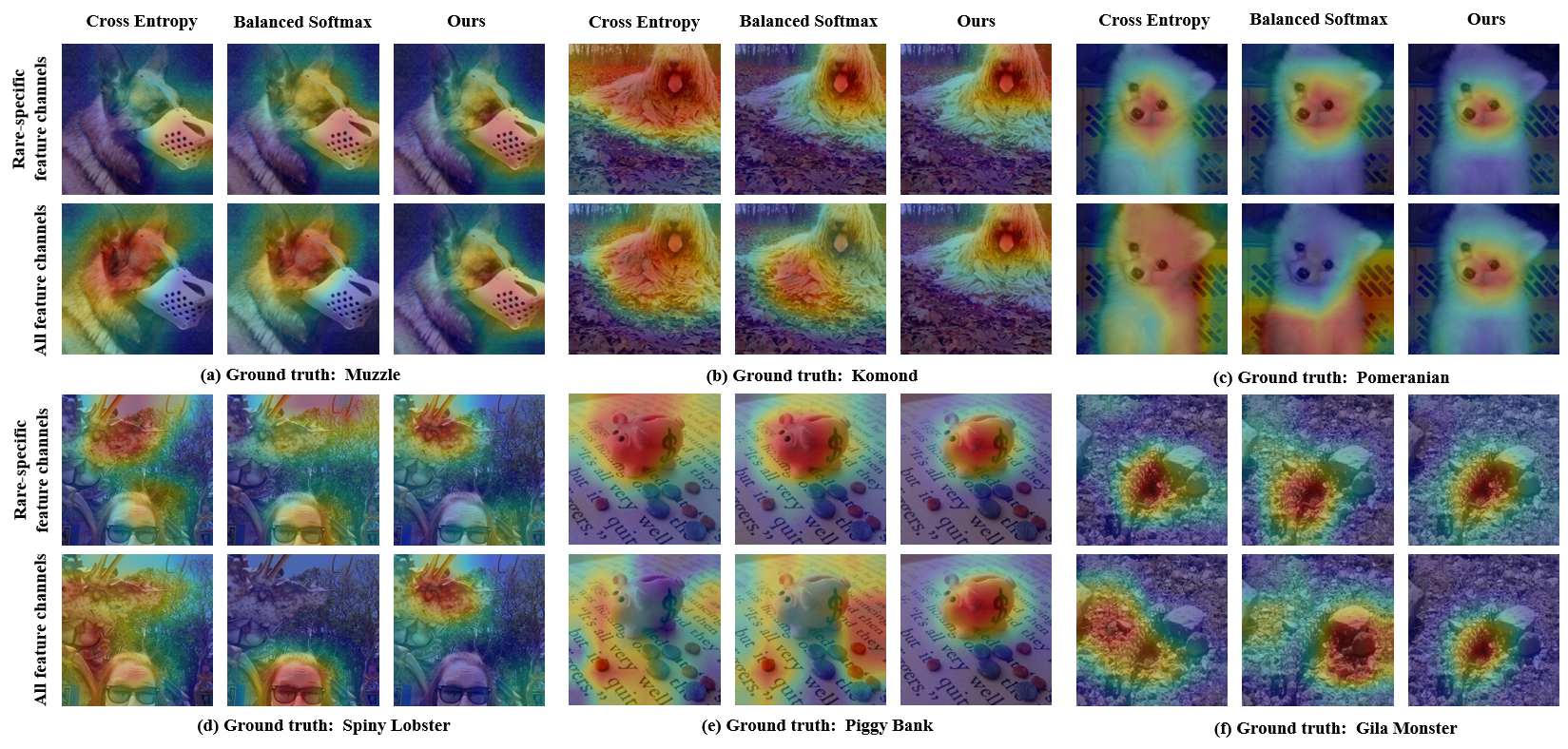}
\vspace{-9mm}
\caption{Visualization of the rare-specific features channels and combination of all the feature channels with the vanilla ResNet-50 trained with cross-entropy loss (\textbf{Cross Entropy}) and Balanced Softmax (\textbf{Balanced Softmax}), and ResNet-50 equipped with NBN trained with cross-entropy loss (\textbf{Ours}). We observe that with the aid of NBN, the model keeps focusing on the class-discriminative regions.}
\label{Fig:specific feature visualization}
\vspace{-3mm}
\end{center}
\end{figure*}

\begin{figure}[t]
\begin{center}
\includegraphics[width=0.8\columnwidth]{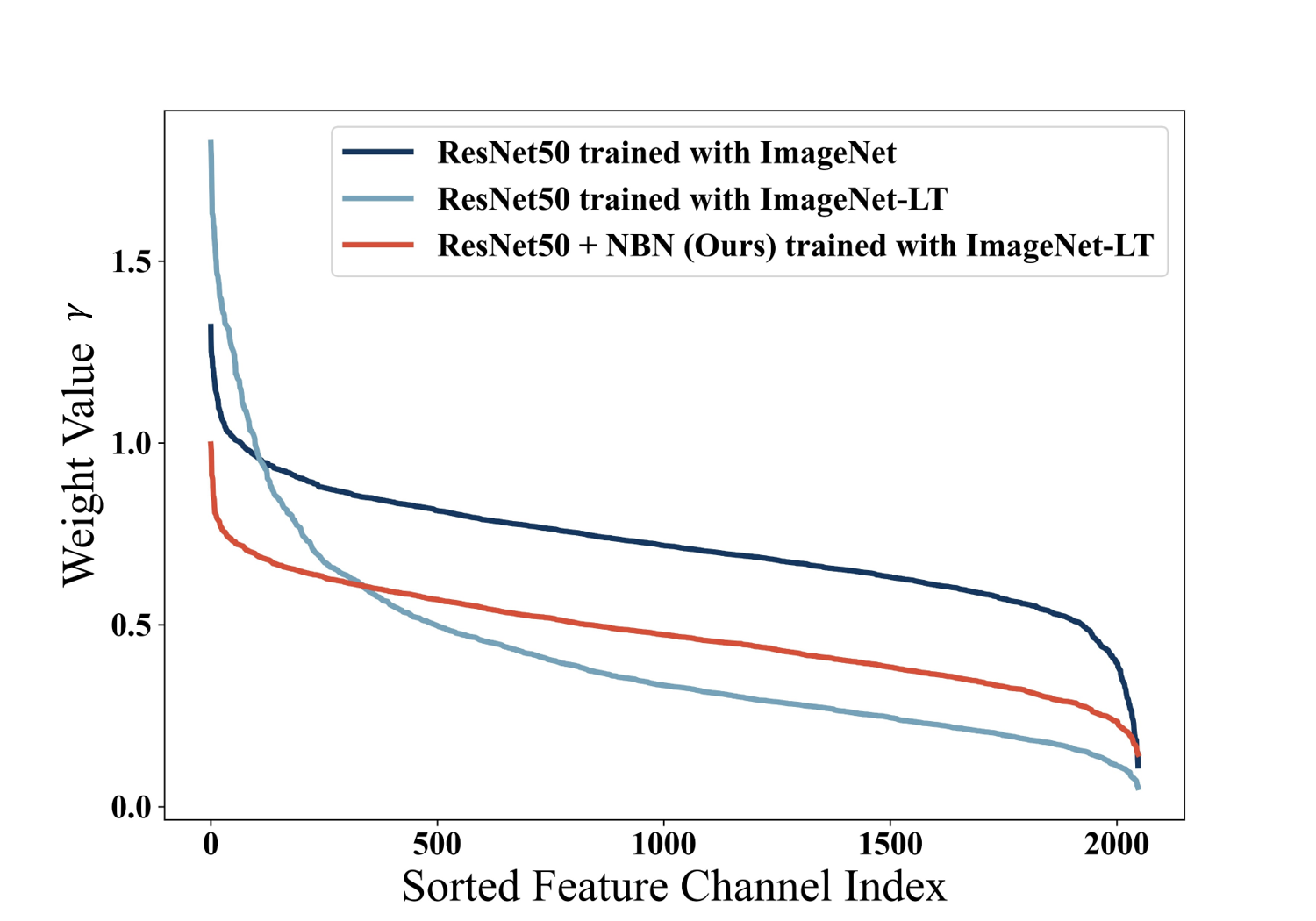}
\vspace{-2mm}
\caption{Comparison of the weight of the last BN layer among the ResNet-50 trained on ImageNet, on ImageNet-LT, and ResNet-50 with NBN on ImageNet-LT. } 
\label{Fig:BN weight visualization}
\end{center}
\vspace{-7mm}
\end{figure}

\begin{table}[t]
\centering
\caption{Ablations for the effect of NBN and logit rectification (LR) with ResNet-50 on ImageNet-LT.}
\resizebox{\linewidth}{!}{
\begin{tabular}{@{}ccccc@{}}
\toprule
   & Baseline & +NBN       & +NBN+LR    \\ \midrule
Cross Entropy & 42.2     & 47.3~\footnotesize{(\textcolor{red}{+5.1})} & 48.8~\footnotesize{(\textcolor{red}{+6.6})} \\
Balanced Softmax & 48.8     & 51.5~\footnotesize{(\textcolor{red}{+2.7})} & 53.0~\footnotesize{(\textcolor{red}{+4.2})} \\ \bottomrule
\end{tabular}}
\vspace{-3mm}
\label{Tab:respective effect}
\end{table}

\noindent \textbf{Effect of NBN and logit rectification (LR).}
In Table~\ref{Tab:respective effect}, we verify the effectiveness of our proposed NBN and logit rectification (LR) module. 
With progressively adding NBN and LR into the architecture, the accuracy of our method is consistently improved, which validates the effectiveness of both NBN and LR.
Specifically, the final performance improvement can be largely attributed to the adoption of NBN. 
As discussed in Sec.~Methods, the LR module may rectify the output statistics and thus improve the accuracy further.

\begin{table}[t]
\centering
\caption{
Comparison of the variance of feature statistics (mean $\mu$ and variance $\sigma$). The experiments are conducted on ImageNet-LT with ResNet-50 backbone. The features that are fed into the classifier are investigated.
}
\resizebox{0.8\linewidth}{!}{
\begin{tabular}{@{}ccc@{}}
\toprule
& Variance of $\mu$ & Variance of $\sigma$ \\ \midrule
Cross Entropy & 0.040                          & 0.152                                \\
Ours     & 0.028                          & 0.074                                \\ \bottomrule
\end{tabular}}
\vspace{-4mm}
\label{Tab: feature statistics}
\end{table}

\begin{figure*}[t]
\centering
\vspace{-3mm}
\includegraphics[scale=0.5]{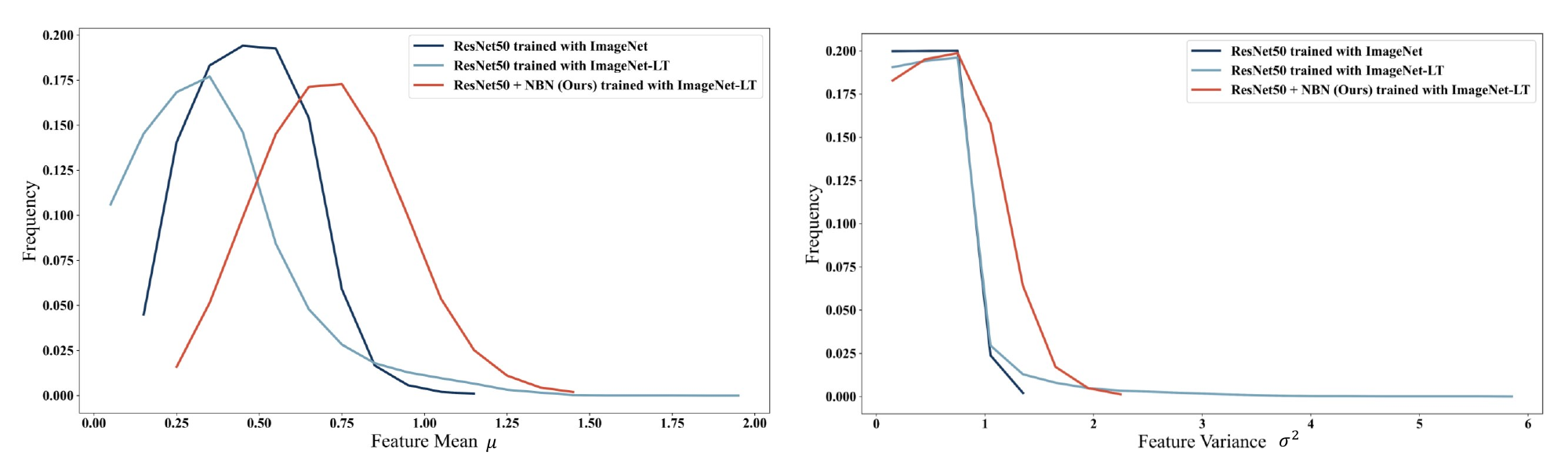}
\vspace{-4mm}
\caption{Histogram of feature statistics across channels. The comparison is among the ResNet-50 trained on ImageNet, ImageNet-LT, and ResNet-50 with NBN on
ImageNet-LT.}
\vspace{-6mm}
\label{fig: histogram}
\end{figure*}

\noindent \textbf{Effect of NBN on rectifying feature bias.}
To demonstrate the rectification effect of NBN, we select three models, 
ResNet-50 trained on ImageNet, 
ResNet-50 trained on ImageNet-LT, and ResNet-50 equipped with NBN trained on ImageNet-LT. 
As shown in Fig.~\ref{Fig:BN weight visualization}, the y-axis represents the Weight value (\emph{i.e.,} $\gamma$) of each channel in the last BN layer of ResNet-50, and the x-axis represents the sorted feature channel index.
The feature channels are sorted in the descending order of $\gamma_k, k\in[0, 2047]$ where $k$ is the channel index. 
It is obvious that the weight curve for the baseline model trained on ImageNet-LT is quite imbalanced. In contrast, the curve corresponding to the one trained on ImageNet is more flat. With the aid of NBN, the skewed curve is flattened and shows similar slope to the one trained on ImageNet. 
We additionally plot the histogram of feature statistics in Fig.~\ref{fig: histogram}. 
We collect the feature embedding of samples from the ImageNet-LT test set and calculate the mean $\mu$ and variance $\sigma$ for each feature channel with ResNet-50 trained on ImageNet~(the dark blue curve), ResNet-50 trained on ImageNet-LT~(the light blue curve), and ResNet-50 inserted with NBN trained on ImageNet-LT~(the red curve) respectively.
It can be seen that there is a long tail in the histogram of ResNet-50 trained on ImageNet-LT compared to the ResNet-50 trained on ImageNet, which means the variance of $\mu$ or $\sigma$ across channels is much larger for ResNet-50 trained on ImageNet-LT than for ResNet-50 trained on ImageNet. With the rectification of NBN, the issue is largely alleviated. 
We also provide the variance of $\mu$ and the variance of $\sigma$ in Table~\ref{Tab: feature statistics}, which are calculated as follows 
\begin{equation}
\label{eq: var(fm)}
\begin{aligned}
\operatorname{Var}(\mu)=\frac{1}{C-1} \sum_{k=1}^C[\mu_k-
\frac{1}{C} \sum_{k=1}^C \mu_k]
\end{aligned}
\end{equation}
\begin{equation}
\label{eq: var(fv)}
\begin{aligned}
\operatorname{Var}(\sigma)=\frac{1}{C-1} \sum_{k=1}^C[\sigma_k-
\frac{1}{C} \sum_{k=1}^C \sigma_k].
\end{aligned}
\end{equation}
As shown in Table~\ref{Tab: feature statistics}, the variance of $\mu$ or $\sigma$ is much larger for the cross-entropy baseline than for our method, which is consistent with what we observe in Fig.~\ref{fig: histogram}.

\noindent \textbf{Comparison between LR and LogN~\cite{zhao2024logit}.}
Both LR and LogN propose to normalize the logits under long-tailed learning scenario. The main difference between the two approaches is that LR is inserted in the network training phase and affects the gradient back-propagation, while LogN is a kind of post-processing method which does not engage in the training.
In Table~\ref{Tab: logn and lr}, we provide the experimental results of ResNext-50 with LR and LogN on ImageNet-LT. We observe that both LogN and LR improve the long-tailed recognition performance. 
Our method (the last row of table) outperforms the results reported by LogN (the third row) by more than 2\%.
By replacing LR with LogN in our solution (see the penultimate row), the accuracy decreases significantly, which shows that LR is more compatible with our designed NBN than LogN.

\begin{table}[h]
\centering
\caption{Comparison between LogN and LR with ResNeXt-50 on ImageNet-LT.}
\label{Tab: logn and lr}
\begin{tabular}{@{}ccccc@{}}
\toprule
Method                    & All  & Tail & Medium & Head \\ \midrule
Cross Entropy             & 45.8 & 9.0  & 39.0   & \textbf{67.2} \\
Balanced Softmax          & 51.1 & 30.2 & 48.1   & 62.3  \\
LogN~\cite{zhao2024logit}    & 51.6 & 35.0 & 50.3   & 59.1 \\
Balanced Softmax + LogN   & 49.6 & 32.7 & 46.6   & 59.3  \\
Balanced Softmax+NBN+LogN & 49.9 & \textbf{37.4} & 46.9   & 58.2 \\
Balanced Softmax+NBN+LR   & \textbf{53.8} & 33.4 & \textbf{50.9}   & 64.6 \\ \bottomrule
\end{tabular}
\end{table}

\subsection{Visualization}
To verify that NBN is effective on enhancing the strength of rare-specific features, we visualize the attention~\cite{zagoruyko2016paying} corresponding to rare-specific feature channels and that corresponding to all feature channels. Specifically, by examining the classifier weights of each category, we select the feature channels which are uniquely important for rare classes.  By aggregating the activation maps of corresponding channels, the image area related to rare-specific features is highlighted, as shown in Fig.~\ref{Fig:specific feature visualization}. By comparing with the attentions of Cross Entropy and Balanced Softmax, we observe that although the rare-specific features have already emerged and captured the discriminative part, their weak strength makes the overall attention unexpectedly focus on irrelevant regions, which explains why the final prediction is incorrect. 
In contrast, our method enables the model to keep focusing on the discriminative features. 
The visualization results validate our assumption and motivation that the rare-specific features are inherently weak, rendering the model biased towards frequent classes.

\section{Conclusion}
In this paper, we observe that under long-tailed scenarios, the learned features may be strongly biased towards the
frequent classes and its distribution may exhibit unexpected imbalance, in the sense that the strengths of discriminative features for rare classes are
weaker than those for frequent classes. To address this issue, we introduce a simple yet
effective strategy, NBN, to explicitly rectify the feature bias
under long-tailed scenarios.  Through extensive experimental results, we demonstrate that our method is plug-and-play and brings consistent improvements compared to previous methods, achieving new state-of-the-arts.

{\small
\bibliographystyle{IEEEtran}
\bibliography{egbib}
}

\section{Biography}
\vspace{-12mm}
\begin{IEEEbiographynophoto}{Yuxiang Bao}
received his M.S. degree from Beihang University in 2024. He is now working as an algorithm engineer at Alibaba Group.
\end{IEEEbiographynophoto}
\vspace{-8mm}
\begin{IEEEbiographynophoto}{Guoliang Kang}
is currently a Professor at Beihang University. 
\end{IEEEbiographynophoto}
\vspace{-8mm}
\begin{IEEEbiographynophoto}{Linlin Yang}
is a Lecturer at the Communication University of China. 
\end{IEEEbiographynophoto}
\vspace{-8mm}
\begin{IEEEbiographynophoto}{Xiaoyue Duan}
is now working as an algorithm engineer at Tencent.
\end{IEEEbiographynophoto}
\vspace{-8mm}
\begin{IEEEbiographynophoto}{Bo Zhao}
is an Associate Professor at Shanghai Jiao Tong University.
\end{IEEEbiographynophoto}
\vspace{-8mm}
\begin{IEEEbiographynophoto}{Baochang Zhang}
is currently a Professor at Beihang University. 
\end{IEEEbiographynophoto}

\vfill

\end{document}